%% file: Paper-1260.tex
\documentclass[runningheads]{llncs}
\usepackage[T1]{fontenc}
\usepackage[pdftex]{graphicx}
\usepackage{verbatim}
\usepackage{amsmath}
\usepackage{amssymb}
\usepackage{amsfonts}
\usepackage{hyperref}

\usepackage[capitalize]{cleveref}
\usepackage{color}

\usepackage{bm}
\renewcommand{\mathbf}[1]{\bm{#1}}
\usepackage{here}
\usepackage{tabularx}
\newcolumntype{R}{>{\raggedright\arraybackslash}X}  
\newcolumntype{C}{>{\centering\arraybackslash}X}     
\newcolumntype{Z}{>{\raggedright\arraybackslash\hsize=10\hsize}X}

\crefname{section}{Sec.}{Secs.}
\Crefname{section}{Section}{Sections}
\Crefname{table}{Table}{Tables}
\crefname{table}{Tab.}{Tabs.}
\begin{document}
\title{Explainable Classifier for Malignant Lymphoma Subtyping via Cell Graph and Image Fusion}
\titlerunning{Explainable Classifier for Malignant Lymphoma Subtyping}

%
%

\author{
Daiki Nishiyama\inst{1,2}\thanks{Corresponding Author}
\and
Hiroaki Miyoshi\inst{3}
\and
Noriaki Hashimoto\inst{2}
\and
Koichi Ohshima\inst{3}
\and
Hidekata Hontani\inst{4}
\and
Ichiro Takeuchi\inst{5,2}
\and
Jun Sakuma\inst{1,2}
}

%
\authorrunning{D. Nishiyama et al.}
%
\institute{
Institute of Science Tokyo, Tokyo, Japan
\\
\email{nishiyama@comp.isct.ac.jp}
\and
RIKEN AIP, Tokyo, Japan
\and
Kurume University, Fukuoka, Japan
\and
Nagoya Institute of Technology, Aichi, Japan
\and
Nagoya University, Aichi, Japan
}
\maketitle              
\begin{abstract}
Lymphoma subtype classification has a direct impact on treatment and outcomes, necessitating models that are both accurate and explainable. This study proposes a novel explainable Multi-Instance Learning (MIL) framework that identifies subtype-specific Regions of Interest (ROIs) from Whole Slide Images (WSIs) while integrating features of cell distribution and image. Our framework simultaneously addresses three objectives: (1) indicating appropriate ROIs for each subtype, (2) explaining the frequency and spatial distribution of characteristic cell types, and (3) reaching accurate subtyping using both cell distribution and image modalities. Our method fuses cell graph and image features extracted for each patch in a WSI by a Mixture-of-Experts-based approach and classifies subtypes within an MIL framework. Experiments on a dataset of 1,233 WSIs demonstrate that our approach achieves state-of-the-art accuracy compared to ten other methods and provides region- and cell-level explanations that align with a pathologist's perspective.
\end{abstract}

\section{Introduction}

\input{docs/1.1background}
\input{docs/1.2.related_works}
\input{docs/1.3.contribution}

\section{Methodology}
\input{docs/3.0.start_method}
\input{docs/3.1.cell_graph}
\input{docs/3.2.model}

\section{Experiments}
\label{sec:experiments}
Requirement for the explainable lymphoma subtyping is to simultaneously address (1) accurate classification of multiple subtypes, (2) identification of subtype-specific ROIs within a WSI, and (3) capture the spatial distribution of characteristic cells present in the ROIs.
In this section, we show that our approach meets the requirement.
\subsection{Experiment Setup}
\input{docs/4.1.exp_setting}
\subsection{Results}
\input{docs/4.2.performance}
\input{docs/4.3.explainability}

\section{Conclusion}
\input{docs/5.conclusion}
\begin{credits}
\subsubsection{\ackname} This work is partly supported by the Japan Science and Technology Agency (JST), CREST JPMJCR21D3, and the Japan Society for the Promotion of Science (JSPS), Grants-in-Aid for Scientific Research 23H00483 and 24KJ1049.
\subsubsection{\discintname}
The authors have no competing interests to declare that are relevant to the content of this article.
\end{credits}
%
%
%
%
\bibliographystyle{splncs04}
\bibliography{Paper-1260}
\end{document}

%% file: docs/1.1background.tex
\label{sec:intro}
\begin{figure}[t]
    \centering
    \includegraphics[width=\linewidth]{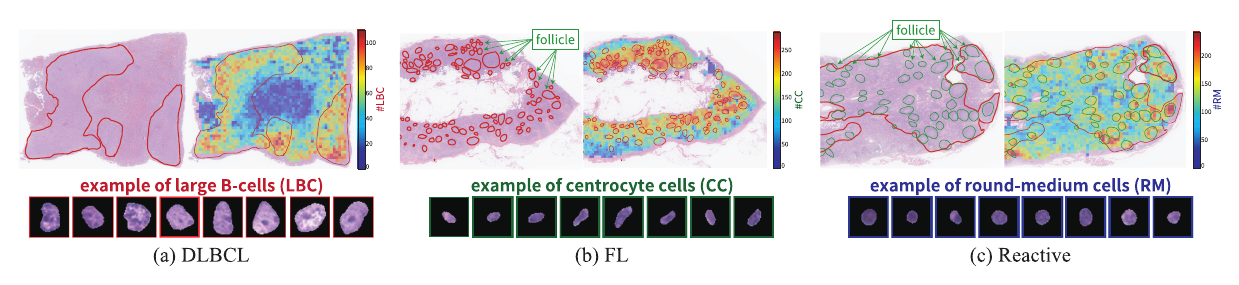}
    \caption{Left to right: DLBCL, FL, and Reactive case, and each subtype's characteristic cells, LBC, CC, and RM. 
    Red lines indicate ROIs, and green and red lines in Reactive and FL, respectively, indicate follicles. 
    Heat maps indicate the spatial distribution of each subtype's number of characteristic cells in a $512\!\times\!512$ pixels at 40x magnification.}
    \label{fig:subtype_overview}
\end{figure}

Subtyping of malignant lymphomas is important because it is essential for determining appropriate treatment strategies and directly affects patient prognosis.
Recent advances in whole slide imaging (WSI) technology \cite{hanna2020whole} have enabled the development of machine learning models capable of classifying malignant lymphoma subtypes from a WSI. 
However, to incorporate a model into actual clinical practice, it must not only achieve high diagnostic accuracy but also provide a reliable explanation for its decisions.

Approaches that mimic the diagnostic processes employed by pathologists are viewed as beneficial for attaining high classification accuracy and reliable explanations. 
Pathologists diagnose lymphoma subtypes by investigating H\&E-stained slide specimens.
They concentrate on local regions of interest (ROIs), which vary for each subtype, and analyze the characteristics, frequency, and spatial distribution of each cell type within these ROIs. During diagnosis, pathologists utilize not only cell-based features but also visual information from pathological images to understand the broader tissue architecture.

As examples, we explain using three clinically important subtypes that we address in this research: diffuse large B-cell lymphoma (DLBCL), follicular lymphoma (FL), and reactive lymphoid hyperplasia (Reactive).
\Cref{fig:subtype_overview} shows, for each subtype, a representative case along with its ROIs, cells corresponding to the subtype, and the distribution of these cells' counts across the WSI.
ROIs of DLBCL show a high frequency of large B-cells (LBCs).
ROIs of FL are inside nearly round structures called \emph{follicles}, where specific cell types such as central cells (CC) overabundantly exist.
ROIs of Reactive extend within and between follicles, and many round-medium cells (RMs) are observed, especially near the follicular boundaries.
As \cref{fig:subtype_overview} indicates, the distribution of cell types characteristic of each subtype is observed only when ROIs specific to that subtype can be properly determined.

Consequently, an explainable malignant lymphoma classification model must be able to (1) identify appropriate ROIs for each subtype, (2) explain the frequency and spatial distribution of characteristic cell types, and (3) achieve high-accuracy subtyping by leveraging both image and cell distribution modalities. 
Cell-level explanations (2) rely on accurately identifying subtype-specific ROIs (1), while the correct determination of ROIs is contingent on understanding cell-type distributions and sustaining overall classification accuracy (3). 
Therefore, our goal is to integrate all three capabilities simultaneously and accomplish a reliable and explainable subtype classification of malignant lymphomas.

%% file: docs/1.2.related_works.tex
\subsection{Related Works}
\label{sec:related_works}
WSI is a gigapixel-scale image that cannot be directly processed by machine learning due to memory constraints.
Multiple Instance Learning (MIL) \cite{maron1997framework} addresses this by partitioning WSIs into a set (called \emph{bag}) of smaller patches (called \emph{instances}).
To indicate ROIs in a WSI, attention-based MILs (ABMILs) \cite{wang2017chestx,ilse2018attention,shao2021transmil} highlight important regions but provide only a single attention score. 
This is insufficient for lymphoma subtyping, where different regions are important for different subtypes. 
AdditiveMIL \cite{javed2022additive} can be a solution to this situation by providing attention scores of instances for each subtype, clarifying the contribution of each instance to the subtyping result.
One limitation of these is that they rely solely on image features without explicitly utilizing cell information.

Cell graphs, where nodes denote cells and edges denote cell adjacencies, effectively capture spatial cell distribution in pathological images.
They can be encoded into feature vectors by graph neural networks to classify subtypes and grades investigated in \cite{abbas2023multi,baranwal2021cgat,gupta2023heterogeneous,jaume2021quantifying,nair2022graph,pati2022hierarchical,sims2022using,zhou2019cgc}.
For instance, HACT-Net \cite{pati2022hierarchical} introduced hierarchical cell graphs to deal with cell- and tissue-level graph structures.
In \cite{jaume2021quantifying}, a post-hoc explainable method was proposed to show morphological attributes of cells important for prediction.
However, these methods are designed for pre-defined and small-scale ROIs rather than WSI-scale, limiting their ability to (1) discover local ROIs specific to the assumed subtype in a WSI and (2) utilize cell graphs for subtyping and explanation across the discovered ROIs.

These limitations are interdependent in malignant lymphoma subtyping and cannot be solved by simply applying existing methods.

%% file: docs/1.3.contribution.tex
\subsection{Contributions}
To resolve these problems, a multi-modal MIL-based model utilizing both image and cell graphs is seen as effective.
Our contributions are summarized as follows:
\begin{itemize}
    \item We propose an explainable classification framework for malignant lymphoma WSIs that indicates localized ROIs through class-wise importance per patch and demonstrates the frequency and spatial distribution of characteristic cell types.
    Experiments confirm that our framework provides region- and cell-level explanations that are well-aligned with a pathologist's view.
    
    \item To achieve high accuracy, we propose a new multi-modal fusion method, Weak-Expert-based Gating Mixture-of-Experts (WEG-MoE), which fuses cell graph and image features for each patch.
    Our method achieves state-of-the-art accuracy compared to ten other methods in classifying three major lymphoma subtypes using our 1,233 WSIs dataset.

    \item According to our investigation, 16 computational pathology studies utilize cell graphs, yet none of them address malignant lymphoma. Our work is, therefore, the first to apply cell graphs specifically to malignant lymphoma.
\end{itemize}

%% file: docs/3.0.start_method.tex
We treat each WSI of lymph node tissue as a \emph{bag} comprising multiple \emph{instances} (patches).
Let $n\in\{1,\ldots, N\}$ be an index of bags (WSIs), $m\in\{1,\ldots, M\}$ an index of instances in a bag, $X_{n,m}$ be the $m$-th image instance in the $n$-th bag, and $\mathbf{y}_n\!\in\!\{0,1\}^K$ be the $n$-th one-hot label over the $K\!=\!3$ subtypes, i.e., DLBCL, FL, and Reactive.
Here, we define $\mathbb{X}_n\!=\!\{X_{n,m}\}_{m=1}^M \!\in\! \mathcal{X}$ as the $n$-th image bag, where $\mathcal{X}$ is a space of image bags.
Then the $n$-th bag is defined by $\mathcal{B}_n\!=\!\bigl(\mathbb{X}_n,\mathbf{y}_n\bigr)$.

%% file: docs/3.1.cell_graph.tex
\subsection{Labeled Cell Graph Construction}
\label{sec:cg}
\begin{figure}[t]
    \centering
    \includegraphics[width=0.92\linewidth]{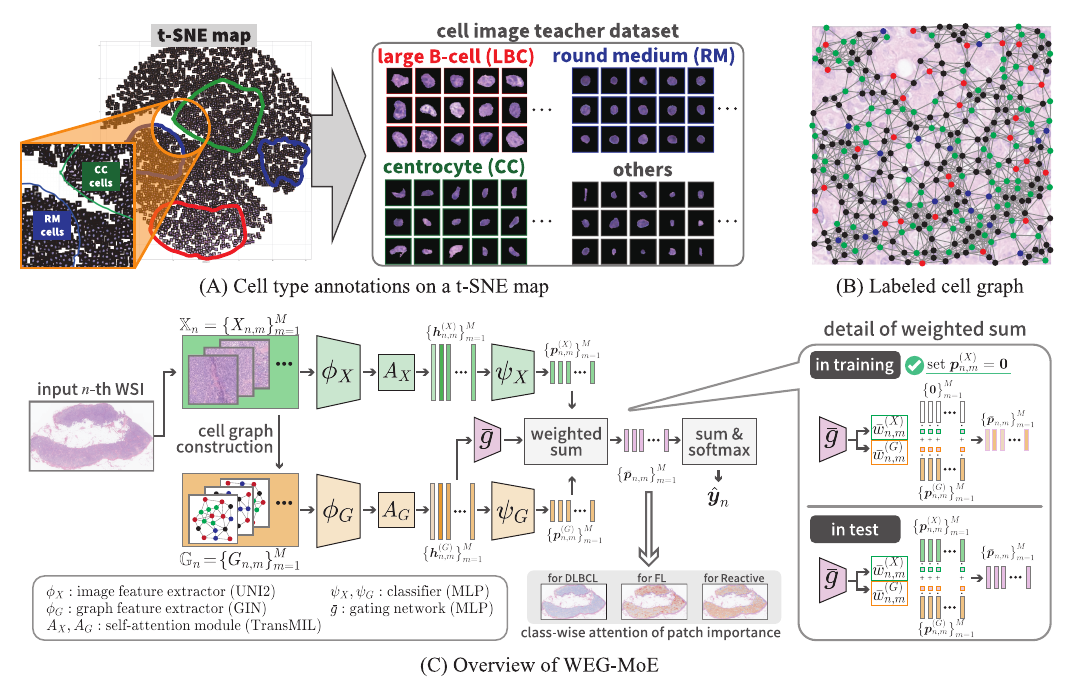}
    \caption{(A) Annotations on a t-SNE map to label cells with LBC, CC, RM, or others. (B) Example of a cell graph to be constructed. The nodes' colors indicate the cell types. (C) An overview of our method, WEG-MoE, for classifying the $n$-th WSI. }
    \label{fig:cell_annot}
\end{figure}
We construct a labeled cell graph $G_{n,m}$ with cells in $X_{n,m}$, where each node label represents a corresponding cell's type.
For cell labeling, we combine cell segmentation using HoVerNet \cite{graham2019hover} pre-trained on CoNSep \cite{graham2019hover} with a cell classifier. 
To train the cell classifier, as \cref{fig:cell_annot}(A) shows, we construct the training dataset by having a pathologist label cell types (LBC, CC, RM, and others) on a 2D t-SNE map. The t-SNE map is created from segmented cells' feature vectors, encoded by ImageNet \cite{he2016deep} pre-trained ResNet34 \cite{deng2009imagenet}.
Identifying LBC, CC, and RM with high recall is crucial for lymphoma subtyping. 
Thus, we employed CAMRI loss 
\cite{nishiyama2023camri}
as the loss function, which enables the classifier to maintain relatively high recall for these three classes while preserving overall accuracy.

Finally, we get the labeled cell graph $G_{n,m} \!=\! (V_{n,m}, E_{n,m}, L_{n,m})$ corresponding to $X_{n,m}$ like \cref{fig:cell_annot}(B), where $V_{n,m}$ is the set of cell indices, $E_{n,m}$ is the set of edges connecting cells within radius $r$ to reflect the spatial density of cells ($r\!=\!60$ in our setting), and $L_{n,m}$ is the set of corresponding cell labels.
Let $\mathbb{G}_n \!=\! \{G_{n,m}\}_{m=1}^M$, then we reformulate the $n$-th WSI bag by $\mathcal{B}_n \!=\! (\mathbb{X}_n, \mathbb{G}_n, \mathbf{y}_n)$.

%% file: docs/3.2.model.tex
\subsection{MoE-based Multimodal MIL}
\label{sec:method_weak_expert_fallback}
As we stated in \cref{sec:intro}, the frequency and spatial distribution of specific cell types are crucial for subtyping lymphoma. 
While labeled cell graphs provide this information, they lack visual elements like cell texture and tissue architecture in pathological images. Then, combining cell graphs with image features is essential for accurate and explainable subtyping.
Therefore, for an input $\mathcal{B}_n\!=\!\bigl(\mathbb{X}_n, \mathbb{G}_n, \mathbf{y}_n\bigr)$, we develop an explainable multi-modal classification model utilizing both graphs and images to achieve high classification performance.

\subsubsection{Pretraining AdditiveMIL for Two Modalities.}
\label{sec:pretrained_experts}
To capture ROIs with different features for each subtype, as we stated in \cref{sec:intro}, we adopt AdditiveMIL \cite{javed2022additive}.
We first introduce AdditiveMIL, where only a single modality (cell graph or image patch) is available.
Let $f_G$ and $f_X$ be MIL models for cell graphs and images, respectively, $\phi_G\!:\!\mathcal{G}\!\to\!\mathbb{R}^d$ be a feature extractor (e.g., GIN \cite{xu2018powerful}) of $f_G$, and $\phi_X\!:\!\mathcal{X}\!\to\!\mathbb{R}^d$ be a feature extractor (e.g., UNI~\cite{chen2024uni}) of $f_X$. 
Given an instance $(X_{n,m}, G_{n,m})$ in a bag $\mathcal{B}_n$, these models produce $d$-dimensional latent features:
\begin{equation}
    \mathbf{h}_{n,m}^{(G)}\!=\!A_G\bigl(\!\phi_G(\!G_{n,1}\!),\! \ldots,\! \phi_G(\!G_{n,M}\!)\!\bigr)_{\!m},
    \mathbf{h}_{n,m}^{(X)}\!=\!A_X\!\bigl(\phi_X\!(X_{n,1}\!),\! \ldots,\! \phi_X\!(\!X_{n,M}\!)\!\bigr)_{\!m},
    \label{eq:latent}
\end{equation}
where $A_G, A_X: \mathbb{R}^{M\times d}\!\rightarrow\!\mathbb{R}^{M\times d}$ are self-attention modules (e.g., TransMIL~\cite{shao2021transmil}).
After each $\mathbf{h}_{n,m}^{(\cdot)}$ is mapped to a $K$-dimensional logit via $\psi_{G}, \psi_{X}: \mathbb{R}^d \!\to\! \mathbb{R}^K$, which are learnable functions (e.g., multilayer perceptrons (MLPs)), finally, we obtain the bag-level class probability:
\begin{align}
\mathbf{p}_{n,m}^{(G)}=\psi_{G}\bigl(\mathbf{h}_{n,m}^{(G)}\bigr),&\quad
\mathbf{p}_{n,m}^{(X)}=\psi_{X}\bigl(\mathbf{h}_{n,m}^{(X)}\bigr),\label{eq:pnm}\\
\hat{\mathbf{y}}_n^{(G)}=\text{softmax}\bigl(\sum\nolimits_{m=1}^M\sigma\bigl(\mathbf{p}_{n,m}^{(G)}\bigr)\bigr),&\quad
\hat{\mathbf{y}}_n^{(X)}=\text{softmax}\bigl(\sum\nolimits_{m=1}^M\sigma\bigl(\mathbf{p}_{n,m}^{(X)}\bigr)\bigr),
\label{eq:bag_prob}
\end{align}
where $\sigma$ is a sigmoid function to stabilize learning in our case (optional), and softmax is a softmax function.
Here, the $j$-th element of $\mathbf{p}_{n,m}^{(\cdot)}$ represents the importance of the $m$-th instance, assuming the prediction result is the $j$-th class.
Because $\hat{\mathbf{y}}_n^{(\cdot)}$ is given as an addition of the class-wise attention score $\mathbf{p}_{n,m}^{(\cdot)}$, the impact of each instance on the classification can be obtained for each class.
$f_G$ and $f_X$ are pre-trained independently using cross-entropy loss $\ell(\hat{\mathbf{y}}_n^{(\cdot)}, \mathbf{y}_n)$.

\subsubsection{Weak-Expert-based Gating MoE.}
To combine cell graph and image modalities, we employ an MoE-based approach.
First, we introduce a naive MoE.
Let $g\!:\!\mathbb{R}^{2d}\!\!\rightarrow\!\![0,1]^2$ be a gating function, and the bag-level prediction by MoE is
\begin{align}
    \hat{\mathbf{y}}_n^{(\mathrm{moe})} &= \text{softmax}\bigl(\sum\nolimits_{m=1}^{M}
    w_{n,m}^{(G)}\;\mathbf{p}_{n,m}^{(G)}+w_{n,m}^{(X)}\;\mathbf{p}_{n,m}^{(X)}\bigr),\label{eq:moe0}\\
    \bigl[w_{n,m}^{(G)},\,w_{n,m}^{(X)}\bigr]&=\mathrm{softmax}
\bigl(
  g(\mathrm{cat}[\mathbf{h}_{n,m}^{(G)}, \mathbf{h}_{n,m}^{(X)} ])
\bigr),
\label{eq:moe}
\end{align}
where $\mathrm{cat}[\cdot,\cdot]$ is a concatenation of two vectors.
Each $w_{n,m}^{(\cdot)}$ in \cref{eq:moe0} and \cref{eq:moe} can be interpreted as how much each modality contributes to the $m$-th instance in the $n$-th bag. 
$g$ allows the model to combine cell graph and image modalities adaptively, considering the contributions of both modalities.

Each modality, i.e., image (ResNet50, UNI2) and graph (GIN), shows decent classification accuracy as shown in \cref{tab:performance}, and the naive MoE simply fuses them without considering modality differences. 
However, the naive MoE actually tends to ignore the graph modality, possibly because the image modality contributes to achieving loss minimization more rapidly.
To better utilize both modalities while preserving explainability, we propose Weak-Expert-based Gating MoE (WEG-MoE), shown in \cref{fig:cell_annot}(C). 
WEG-MoE gates each modality using only the weaker modality (i.e., graph), which typically shows lower accuracy.
Let $\bar{g}\!:\!\mathbb{R}^{d}\!\!\rightarrow\!\![0,1]^2$ be WEG-MoE's gating function. 
The bag-level prediction by WEG-MoE is
\begin{align}
    \hat{\mathbf{y}}_n^{(\mathrm{weg})} &=\text{softmax}\bigl(\sum\nolimits_{m=1}^{M}
    \bar{w}_{n,m}^{(G)}\;\mathbf{p}_{n,m}^{(G)}+\bar{w}_{n,m}^{(X)}\;\mathbf{p}_{n,m}^{(X)}\bigr),\label{eq:weg}\\
    \text{where}\quad\bigl[\bar{w}_{n,m}^{(G)}, \bar{w}_{n,m}^{(X)}\bigr]&=\mathrm{softmax}
    \bigl(
        \bar{g}\bigl(\mathbf{h}_{n,m}^{(G)}\bigr)
    \bigr).
\end{align}
During training, after independently pre-training each modality model $f_G$ and $f_X$, we force $\mathbf{p}_{n,m}^{(X)}=\boldsymbol{0}$ and optimize only $\bar{g}$'s parameters.
In this way, WEG-MoE determines whether the graph modality is useful for each patch and uses the image modality as necessary.

%% file: docs/4.1.exp_setting.tex
Because public datasets have only a single subtype label \cite{tcga_dlbc,fernandez-pol2022he} or ROIs without full WSIs \cite{kaggle_lymphoma}, these datasets are unsuitable for evaluation based on the requirement as mentioned above.
Then, we constructed a private dataset comprising 1,233 lymphoma WSIs diagnosed at Kurume University, with labels of DLBCL, FL, and Reactive subtypes (411 per subtype).
For training the cell classifier, 237,544 cell images were carefully selected to prevent test data leakage.
Annotation of cell labels by a pathologist was completed in a few minutes using the annotation method introduced in \cref{sec:cg}.
Each patch was clipped at 512$\!\times\!$512 pixels from a non-background area and selected if it had at least 100 cells.

We compared the performance of our WEG-MoE with several baselines.
Unimodal baselines included ResNet50 \cite{he2016deep}, UNI2 \cite{chen2024uni}, GIN \cite{xu2018powerful}, and HACT-Net \cite{pati2022hierarchical}.
Multi-modal baselines included concat \cite{mobadersany2018predicting}, Pathomic Fusion \cite{chen2020pathomic}, MCAT \cite{chen2021multimodal}, mutual attention \cite{cai2023multimodal}, and MoE \cite{hashimoto2024multimodal}.
To adapt all cases to MIL, AdditiveMIL \cite{javed2022additive} and TransMIL \cite{shao2021transmil} served as $(f_X,\! A_X\!)$ and $(f_G,\! A_G\!)$, respectively. In AdditiveMIL, $\psi_X$ and $\psi_G$ were three fully connected (FC) layers with ReLU activation.
In UNI2, an FC layer followed UNI2's feature extractor, and only this FC layer was trained.
In GIN, the feature extractor consisted of four GIN layers.
In HACT-Net, we used deep features as node features, following \cite{pati2022hierarchical}.
In multi-modal baselines, UNI2 was $\phi_X$, and GIN was $\phi_G$, and $\mathbf{h}_{n,m}^{(X)}$ and $\mathbf{h}_{n,m}^{(G)}$ with latent dimension $d\!=\!256$ were fused.
In Pathomic Fusion, we reduced the feature dimension to 48 before the direct product.
In MCAT, we employed two flows to handle its directional attention flow: graph-to-image (g2i) and image-to-graph (i2g).
In MoE and WEG-MoE, $g$ and $\bar{g}$ were three FC layers with ReLU activation.

We used ten NVIDIA A100 80GB GPUs. In WEG-MoE, we processed ten WSIs simultaneously in parallel, taking approximately one hour in total from WSI scanning to test prediction.
Our implementation is publicly available\footnote{\url{https://github.com/mdl-lab/Explainable-Malignant-Lymphoma-Classifier}}.

%% file: docs/4.2.performance.tex
\begin{table}[t]
    \centering
    \caption{Mean and standard deviation across five cross-validation sets: accuracy, each class AUC, and mean AUC.
    The table is divided into image modality, graph modality, and multi-modality sections by lines from the top. Best performance is in bold.}
    \label{tab:performance}
    {
    \fontsize{8pt}{9.6pt}\selectfont
    \begin{tabularx}{\textwidth}{l|C|C|C|c|C}
        \hline
        Method & Accuracy & DLBCL AUC & FL AUC & Reactive AUC & AUC mean \\
        \hline
        \hline
        ResNet50\cite{he2016deep} & $0.842\pm0.032$ & $0.956\pm0.022$ & $0.870\pm0.059$ & $0.973\pm0.016$ & $0.933\pm0.045$ \\
        UNI2\cite{chen2024uni}& $0.904\pm0.014$ & $0.980\pm0.010$ & $0.948\pm0.015$ & $0.988\pm0.010$ & $0.972\pm0.017$ \\
        \hline
        GIN\cite{xu2018powerful}&$0.829\pm0.019$ & $0.964\pm0.012$ & $0.807\pm0.047$ & $0.916\pm0.025$ & $0.896\pm0.065$\\
        HACT-Net\cite{pati2022hierarchical}&$0.761\pm0.077$ & $0.957\pm0.022$ & $0.856\pm0.045$ & $0.943\pm0.023$ & $0.919\pm0.045$\\
        \hline
        concat\cite{mobadersany2018predicting} & $0.907\pm0.012$ & $0.980\pm0.011$ & $0.954\pm0.002$ & $\boldsymbol{0.989\!\pm\!0.008}$ & $0.975\pm0.015$ \\
        Pathomic Fusion\cite{chen2020pathomic}& $0.902\pm0.015$ & $0.975\pm0.017$ & $0.949\pm0.016$ & $0.986\pm0.012$ & $0.970\pm0.015$ \\
        MCAT g2i\cite{chen2021multimodal}&$0.901\pm0.011$ & $0.973\pm0.010$ & $0.953\pm0.013$ & $0.988\pm0.004$ & $0.971\pm0.014$\\
        MCAT i2g\cite{chen2021multimodal}&$0.881\pm0.019$ & $0.975\pm0.013$ & $0.951\pm0.015$ & $0.986\pm0.006$ & $0.971\pm0.015$ \\
        mutual attention\cite{cai2023multimodal}&$0.906\pm0.018$ & $0.972\pm0.017$ & $0.956\pm0.008$ & $0.987\pm0.007$ & $0.972\pm0.013$\\
        MoE\cite{hashimoto2024multimodal}& $ 0.907\pm0.012$ & $0.977\pm0.013$ & $0.951\pm0.016$ & $0.988\pm0.010$ & $0.972\pm0.015$  \\
        WEG-MoE (ours)& $\boldsymbol{0.911\!\pm\!0.011}$ & $\boldsymbol{0.983\!\pm\!0.009}$ & $\boldsymbol{0.961\!\pm\!0.011}$ & $ 0.988\pm0.010 $ & $\boldsymbol{0.977\!\pm\!0.012}$ \\
        \hline
    \end{tabularx}
    }
\end{table}
\Cref{tab:performance} shows classification performance.
We can see that WEG-MoE outperforms the others in classification performance.
Especially since WEG-MoE outperforms MoE, we can confirm the effectiveness of WEG-MoE's gating strategy.

%% file: docs/4.3.explainability.tex
\begin{figure}[t]
    \centering
    \includegraphics[width=0.9\linewidth]{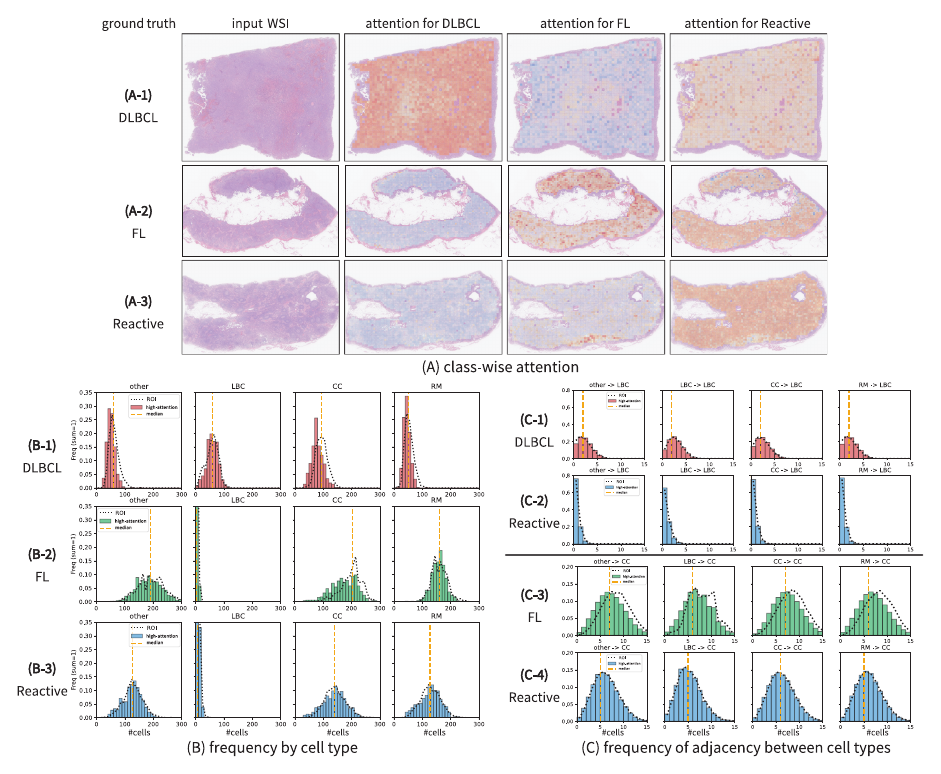}
    \caption{(A) Class-wise attention, where higher attention is red, and lower is blue.
    (B) Frequency by cell types. (C) Frequency of adjacency between cell types.
    In (B) and (C), the black dotted lines present the results computed with the ROI supervised by a pathologist, which can be interpreted as the ground truth, and the orange dotted lines present the median of the distribution.
    In (C), for example, ``LBC$\to$CC'' indicates the number of CCs that are connected by edges to LBC.
    Each data in (B) and (C) is from the top 25\% of instance-level class-wise attention scores for the correct class.}
    \label{fig:exp_result}
\end{figure}

\Cref{fig:exp_result} shows explainability results with a representative case:
(A) class-wise attention of each subtype,
(B) frequency by cell type, and (C) frequency of cell adjacency.
(B) and (C) are computed from high-attention regions (top 25\%).
Dotted lines show distribution within pathologist-supervised ROIs.
Below, we contrast what the pathologist expects with explanations by our framework.

\emph{DLBCL.}
\textbf{Expectation:}
DLBCL has spread lesions with increased and diffusely distributed LBCs.
\textbf{Attention:}
\Cref{fig:exp_result}(A-1) shows higher attention for DLBCL overall than other subtypes, reflecting that LBCs are diffusely distributed overall. 
\textbf{Cell frequency:} 
\Cref{fig:exp_result}(B-1) indicates increased LBCs compared to other subtypes.
\textbf{Cell distribution:}
LBCs in DLBCL are adjacent to a greater variety of cells than in Reactive (\cref{fig:exp_result}(C-1) vs (C-2)).
\textbf{Validity:}
These results are consistent with the expectation, and the distribution of LBC is close to it in the supervised ROI, so they are aligned with the pathologist's view.

\emph{FL.}
\textbf{Expectation:}
FL lesions are follicles with densely packed cells and numerous CCs.
\textbf{Attention:}
In specific regions, \cref{fig:exp_result}(A-2) shows higher attention for FL, while it is lower for Reactive and DLBCL.
\textbf{Cell frequency:} 
\Cref{fig:exp_result}(B-2) indicates increased CCs compared to other subtypes.
\textbf{Cell distribution:}
The number of CCs adjacent to various cells in FL (\cref{fig:exp_result}(C-3)) exceeds those in Reactive (\cref{fig:exp_result}(C-4)).
\textbf{Validity:}
These results indicate that the follicles show attention and have the characteristics of FL rather than Reactive, so they are aligned with the pathologist's view.

\emph{Reactive.}
\textbf{Expectation:}
Reactive tissue shows normal follicular structure and inter-follicular tissue throughout, with RM often at follicle boundaries.
\textbf{Attention:}
\Cref{fig:exp_result}(A-3) shows high attention for Reactive overall, with locally high FL attention in the lower part.
\textbf{Cell frequency:} 
Reactive shows fewer LBCs than DLBCL (\cref{fig:exp_result}(B-3) vs (B-1)) and fewer CCs, RMs, and other cells than FL (\cref{fig:exp_result}(B-3) vs (B-2)).
\textbf{Cell distribution:}
Reactive shows fewer cell adjacencies for both LBCs and CCs compared to DLBCL and FL, respectively.
\textbf{Validity:}
Areas with high FL attention appear as FL to pathologists, while other regions display Reactive characteristics rather than malignant patterns, validating the model's attention allocation.
Due to the Reactive ROIs covering areas both within and outside follicles, it can not be observed that the RM frequency is high at follicular borders.
When comparing cell distributions, the Reactive case displays non-malignant patterns distinct from DLBCL and FL. These patterns closely match the distribution in pathologist-identified ROIs, confirming the overall appropriateness of the model's interpretations.

%% file: docs/5.conclusion.tex
We proposed an explainable multi-modal MIL framework for the subtyping of malignant lymphoma. 
This framework can explain not only localized ROIs through class-wise attention but also the frequency and spatial distribution of characteristic cell types based on the labeled cell graph.
Our experiments confirmed the appropriateness of these explanations based on the pathologist's assessment and showed state-of-the-art classification performance.

%% file: Paper-1260.bib
@inproceedings{he2016deep,
    title = {Deep residual learning for image recognition},
    author = {He, Kaiming and Zhang, Xiangyu and Ren, Shaoqing and Sun, Jian},
    booktitle = {Proceedings of the IEEE conference on computer vision and pattern recognition},
    pages = {770--778},
    year = {2016}
}

@article{graham2019hover,
    title = {Hover-net: Simultaneous segmentation and classification of nuclei in multi-tissue histology images},
    author = {Graham, Simon and Vu, Quoc Dang and Raza, Shan E Ahmed and Azam, Ayesha and Tsang, Yee Wah and Kwak, Jin Tae and Rajpoot, Nasir},
    journal = {Medical Image Analysis},
    volume = {58},
    pages = {101563},
    year = {2019},
    publisher = {Elsevier}
}

@inproceedings{ilse2018attention,
    title = {Attention-based deep multiple instance learning},
    author = {Ilse, Maximilian and Tomczak, Jakub and Welling, Max},
    booktitle = {International conference on machine learning},
    pages = {2127--2136},
    year = {2018},
    organization = {PMLR}
}

@article{chen2020pathomic,
    title = {Pathomic fusion: an integrated framework for fusing histopathology and genomic features for cancer diagnosis and prognosis},
    author = {Chen, Richard J and Lu, Ming Y and Wang, Jingwen and Williamson, Drew FK and Rodig, Scott J and Lindeman, Neal I and Mahmood, Faisal},
    journal = {IEEE Transactions on Medical Imaging},
    year = {2020},
    publisher = {IEEE}
}

@inproceedings{zhou2019cgc,
  title={CGC-Net: Cell Graph Convolutional Network for Grading of Colorectal Cancer Histology Images},
  author={Zhou, Yanning and Graham, Simon and Koohbanani, Navid Alemi and Shaban, Muhammad and Heng, Pheng-Ann and Rajpoot, Nasir},
  booktitle={The IEEE International Conference on Computer Vision (ICCV) Workshops}, 
  year={2019}
}

@article{pati2022hierarchical,
title = {Hierarchical graph representations in digital pathology},
journal = {Medical Image Analysis},
volume = {75},
pages = {102264},
year = {2022},
issn = {1361-8415},
author = {Pushpak Pati and Guillaume Jaume and Antonio Foncubierta-Rodríguez and Florinda Feroce and Anna Maria Anniciello and Giosue Scognamiglio and Nadia Brancati and Maryse Fiche and Estelle Dubruc and Daniel Riccio and Maurizio {Di Bonito} and Giuseppe {De Pietro} and Gerardo Botti and Jean-Philippe Thiran and Maria Frucci and Orcun Goksel and Maria Gabrani}
}

@article{shao2021transmil,
    title = {Transmil: Transformer based correlated multiple instance learning for whole slide image classification},
    author = {Shao, Zhuchen and Bian, Hao and Chen, Yang and Wang, Yifeng and Zhang, Jian and Ji, Xiangyang and others},
    journal = {Advances in Neural Information Processing Systems},
    volume = {34},
    pages = {2136--2147},
    year = {2021}
}

@article{javed2022additive,
    title = {Additive MIL: Intrinsically Interpretable Multiple Instance Learning for Pathology},
    author = {Javed, Syed Ashar and Juyal, Dinkar and Padigela, Harshith and Taylor-Weiner, Amaro and Yu, Limin and Prakash, Aaditya},
    journal = {Advances in Neural Information Processing Systems},
    volume = {35},
    pages = {20689--20702},
    year = {2022}
}

@inproceedings{jaume2021quantifying,
    title = {Quantifying explainers of graph neural networks in computational pathology},
    author = {Jaume, Guillaume and Pati, Pushpak and Bozorgtabar, Behzad and Foncubierta, Antonio and Anniciello, Anna Maria and Feroce, Florinda and Rau, Tilman and Thiran, Jean-Philippe and Gabrani, Maria and Goksel, Orcun},
    booktitle = {Proceedings of the IEEE/CVF conference on computer vision and pattern recognition},
    pages = {8106--8116},
    year = {2021}
}

@article{chen2024uni,
  title={Towards a General-Purpose Foundation Model for Computational Pathology},
  author={Chen, Richard J and Ding, Tong and Lu, Ming Y and Williamson, Drew FK and Jaume, Guillaume and Chen, Bowen and Zhang, Andrew and Shao, Daniel and Song, Andrew H and Shaban, Muhammad and others},
  journal={Nature Medicine},
  publisher={Nature Publishing Group},
  year={2024}
}

@article{nishiyama2023camri,
  title={CAMRI Loss: Improving the Recall of a Specific Class without Sacrificing Accuracy},
  author={Nishiyama, Daiki and Fukuchi, Kazuto and Akimoto, Youhei and Sakuma, Jun},
  journal={IEICE TRANSACTIONS on Information and Systems},
  volume={106},
  number={4},
  pages={523--537},
  year={2023},
  publisher={The Institute of Electronics, Information and Communication Engineers}
}

@inproceedings{deng2009imagenet,
  title={Imagenet: A large-scale hierarchical image database},
  author={Deng, Jia and Dong, Wei and Socher, Richard and Li, Li-Jia and Li, Kai and Fei-Fei, Li},
  booktitle={2009 IEEE conference on computer vision and pattern recognition},
  pages={248--255},
  year={2009},
  organization={Ieee}
}

@article{maron1997framework,
  title={A framework for multiple-instance learning},
  author={Maron, Oded and Lozano-P{\'e}rez, Tom{\'a}s},
  journal={Advances in neural information processing systems},
  volume={10},
  year={1997}
}

@inproceedings{
    xu2018powerful,
    title={How Powerful are Graph Neural Networks?},
    author={Keyulu Xu and Weihua Hu and Jure Leskovec and Stefanie Jegelka},
    booktitle={International Conference on Learning Representations},
    year={2019}
}

@inproceedings{wang2017chestx,
  title={Chestx-ray8: Hospital-scale chest x-ray database and benchmarks on weakly-supervised classification and localization of common thorax diseases},
  author={Wang, Xiaosong and Peng, Yifan and Lu, Le and Lu, Zhiyong and Bagheri, Mohammadhadi and Summers, Ronald M},
  booktitle={Proceedings of the IEEE conference on computer vision and pattern recognition},
  pages={2097--2106},
  year={2017}
}

@article{hanna2020whole,
  title={Whole slide imaging: technology and applications},
  author={Hanna, Matthew G and Parwani, Anil and Sirintrapun, Sahussapont Joseph},
  journal={Advances in Anatomic Pathology},
  volume={27},
  number={4},
  pages={251--259},
  year={2020},
  publisher={LWW}
}

@article{abbas2023multi, title={Multi-cell type and multi-level graph aggregation network for cancer grading in pathology images}, author={Abbas, Syed Farhan and Le Vuong, Trinh Thi and Kim, Kyungeun and Song, Boram and Kwak, Jin Tae}, journal={Medical Image Analysis}, volume={90}, pages={102936}, year={2023}, publisher={Elsevier} }

@InProceedings{gupta2023heterogeneous,
author="Krishna, Akhila
and Gupta, Ravi Kant
and Kurian, Nikhil Cherian
and Jeevan, Pranav
and Sethi, Amit",
editor="Ahmadi, Seyed-Ahmad
and Pereira, S{\'e}rgio",
title="Heterogeneous Graphs Model Spatial Relationship Between Biological Entities for Breast Cancer Diagnosis",
booktitle="Graphs in Biomedical Image Analysis, and Overlapped Cell on Tissue Dataset for Histopathology",
year="2024",
publisher="Springer Nature Switzerland",
address="Cham",
pages="97--106",
isbn="978-3-031-55088-1"
}

@inproceedings{sims2022using,
  title={Using hierarchically connected nodes and multiple gnn message passing steps to increase the contextual information in cell-graph classification},
  author={Sims, Joe and Grabsch, Heike I and Magee, Derek},
  booktitle={MICCAI Workshop on Imaging Systems for GI Endoscopy},
  pages={99--107},
  year={2022},
  organization={Springer}
}

@article{baranwal2021cgat,
  title={Cgat: Cell graph attention network for grading of pancreatic disease histology images},
  author={Baranwal, Mayank and Krishnan, Santhoshi and Oneka, Morgan and Frankel, Timothy and Rao, Arvind},
  journal={Frontiers in Immunology},
  volume={12},
  pages={727610},
  year={2021},
  publisher={Frontiers Media SA}
}

@article{nair2022graph,
  title={A graph neural network framework for mapping histological topology in oral mucosal tissue},
  author={Nair, Aravind and Arvidsson, Helena and Gatica V, Jorge E and Tudzarovski, Nikolce and Meinke, Karl and Sugars, Rachael V},
  journal={BMC bioinformatics},
  volume={23},
  number={1},
  pages={506},
  year={2022},
  publisher={Springer}
}

@article{mobadersany2018predicting,
  title={Predicting cancer outcomes from histology and genomics using convolutional networks},
  author={Mobadersany, Pooya and Yousefi, Safoora and Amgad, Mohamed and Gutman, David A and Barnholtz-Sloan, Jill S and Vel{\'a}zquez Vega, Jos{\'e} E and Brat, Daniel J and Cooper, Lee AD},
  journal={Proceedings of the National Academy of Sciences},
  volume={115},
  number={13},
  pages={E2970--E2979},
  year={2018},
  publisher={National Academy of Sciences}
}

@inproceedings{chen2021multimodal,
  title={Multimodal co-attention transformer for survival prediction in gigapixel whole slide images},
  author={Chen, Richard J and Lu, Ming Y and Weng, Wei-Hung and Chen, Tiffany Y and Williamson, Drew FK and Manz, Trevor and Shady, Maha and Mahmood, Faisal},
  booktitle={Proceedings of the IEEE/CVF international conference on computer vision},
  pages={4015--4025},
  year={2021}
}

@article{hashimoto2024multimodal,
  title={Multimodal gated mixture of experts using whole slide image and flow cytometry for multiple instance learning classification of lymphoma},
  author={Hashimoto, Noriaki and Hanada, Hiroyuki and Miyoshi, Hiroaki and Nagaishi, Miharu and Sato, Kensaku and Hontani, Hidekata and Ohshima, Koichi and Takeuchi, Ichiro},
  journal={Journal of Pathology Informatics},
  volume={15},
  pages={100359},
  year={2024},
  publisher={Elsevier}
}

@article{cai2023multimodal,
  title={A multimodal transformer to fuse images and metadata for skin disease classification},
  author={Cai, Gan and Zhu, Yu and Wu, Yue and Jiang, Xiaoben and Ye, Jiongyao and Yang, Dawei},
  journal={The Visual Computer},
  volume={39},
  number={7},
  pages={2781--2793},
  year={2023},
  publisher={Springer}
}

@misc{fernandez-pol2022he,
  author = {Fernandez-Pol, S. and Natkunam, Y. and Vrabac, D. and Rojansky, R. and Advani, R. and Rajpurkar, P. and S and Ng, Andrew Y.},
  title = {H\&E and immunohistochemical stain images of 209 cases of diffuse large B-cell lymphoma linked with cytogenetic features and clinical outcomes (Version 1) [Data set]},
  year = {2022},
  publisher = {The Cancer Imaging Archive},
  doi = {10.7937/NVA3-N783}
}

@misc{tcga_dlbc,
  author = {{The Cancer Genome Atlas Research Network}},
  title = {The Cancer Genome Atlas Lymphoid Neoplasm Diffuse Large B-cell Lymphoma (TCGA-DLBC)},
  year = {2016},
  publisher = {National Cancer Institute},
  howpublished = {Genomic Data Commons Data Portal},
  url = {https://portal.gdc.cancer.gov/projects/TCGA-DLBC}
}

@article{kaggle_lymphoma,
author = {Orlov, Nikita and Chen, Wayne and Eckley, David and Macura, Tomasz and Shamir, Lior and Jaffe, Elaine and Goldberg, Ilya},
year = {2010},
month = {07},
pages = {1003-13},
title = {Automatic Classification of Lymphoma Images With Transform-Based Global Features},
volume = {14},
journal = {IEEE transactions on information technology in biomedicine : a publication of the IEEE Engineering in Medicine and Biology Society},
doi = {10.1109/TITB.2010.2050695}
}
